\pdfoutput=1

\documentclass[11pt]{article}

\usepackage[preprint]{acl}

\usepackage{times}
\usepackage{latexsym}

\usepackage[T1]{fontenc}

\usepackage[utf8]{inputenc}

\usepackage{microtype}

\usepackage{inconsolata}

\usepackage{inconsolata}
\usepackage{amsmath,amsfonts}
\usepackage{algorithmic}
\usepackage{algorithm}
\usepackage{array}
\usepackage[caption=false,font=normalsize,labelfont=sf,textfont=sf]{subfig}
\usepackage{textcomp}
\usepackage{stfloats}
\usepackage{url}
\usepackage{verbatim}
\usepackage{graphicx}
\usepackage{cite}
\usepackage{xcolor}
\usepackage{threeparttable}
\usepackage{makecell}
\usepackage{multirow}
\usepackage{orcidlink}
\usepackage{booktabs} 
\usepackage{enumitem}

\usepackage{tabularx}
\usepackage{hyperref}
\usepackage{xurl}

\newcounter{bxcomm}
\definecolor{aqua}{rgb}{0.00,0.67,0.80}

\newcommand{\eg}{\textit{e}.\textit{g}.}

\newcommand{\vf}{\textsc{VenusFactory}}

\title{\vf: A Unified Platform for Protein Engineering Data Retrieval and Language Model Fine-Tuning}

\author{
  \bf Yang Tan$^{1,2,3,}$\thanks{~~Equal contribution and this work was done during the internship at Shanghai Artificial Intelligence Laboratory.}, 
  \bf Chen Liu$^{3,*}$, 
  \bf Jingyuan Gao$^{1,*}$, 
  \bf Banghao Wu$^1$, 
  \bf Mingchen Li$^{1,2,3}$,\\ 
  \bf Ruilin Wang$^3$, 
  \bf Lingrong Zhang$^{1}$, 
  \bf Huiqun Yu$^3$, 
  \bf Guisheng Fan$^{3}$, 
  \bf Liang Hong$^{1,2}$, 
  \bf Bingxin Zhou$^{1,}$\thanks{~~Corresponding author (bingxin.zhou@sjtu.edu.cn).} \\
  $^1$ Shanghai Jiao Tong University, China \\
  $^2$ Shanghai Artificial Intelligence Laboratory, China \\
  $^3$ East China University of Science and Technology, China\\
}

\begin{document}
\maketitle
\begin{abstract}
Natural language processing (NLP) has significantly influenced scientific domains beyond human language, including protein engineering, where pre-trained protein language models (PLMs) have demonstrated remarkable success. However, interdisciplinary adoption remains limited due to challenges in data collection, task benchmarking, and application. This work presents \vf, a versatile engine that integrates biological data retrieval, standardized task benchmarking, and modular fine-tuning of PLMs. \vf~supports both computer science and biology communities with choices of both a command-line execution and a Gradio-based no-code interface, integrating $40+$ protein-related datasets and $40+$ popular PLMs. All implementations are open-sourced on \url{https://github.com/tyang816/VenusFactory}.
\end{abstract}

\section{Introduction}
Discrete tokens provide a natural representation of data across various fields, including human language, amino acid sequences, and molecular structures \citep{brown2020gpt3,guo2025deepseek-r1}. The recent success of natural language processing and large language models has introduced novel solutions to fundamental scientific and engineering challenges \citep{pan2023llm_chem,zhou2024mlife}. In enzyme engineering, pre-trained protein language models (PLMs) have been developed to analyze and extract hidden amino acid interactions and evolutionary features from protein sequences \citep{meier2021esm1v,rives2021esm1b,tan2023protssn,li2024prosst}. The growing interest in AI-driven scientific research in protein engineering has led to the development of many open-source PLMs for both the computer science and computational biology communities. For example, ESM2-650M \citep{lin2023esm2}, arguably the most popular and powerful sequence-encoding PLM, has over one million downloads per month from HuggingFace\footnote{\url{https://huggingface.co/facebook/esm2_t33_650M_UR50D}}. Meanwhile, by integrating task-specific labeled data and predictive modules, these models facilitate downstream tasks such as sequence generation, catalytic activity enhancement, function prediction, and properties assessment, thereby advancing enzyme production and application \citep{madani2023progen,zhou2024cpdiffusion,kang2024vhh}.

\begin{figure*}
    \centering
    \includegraphics[width=\textwidth]{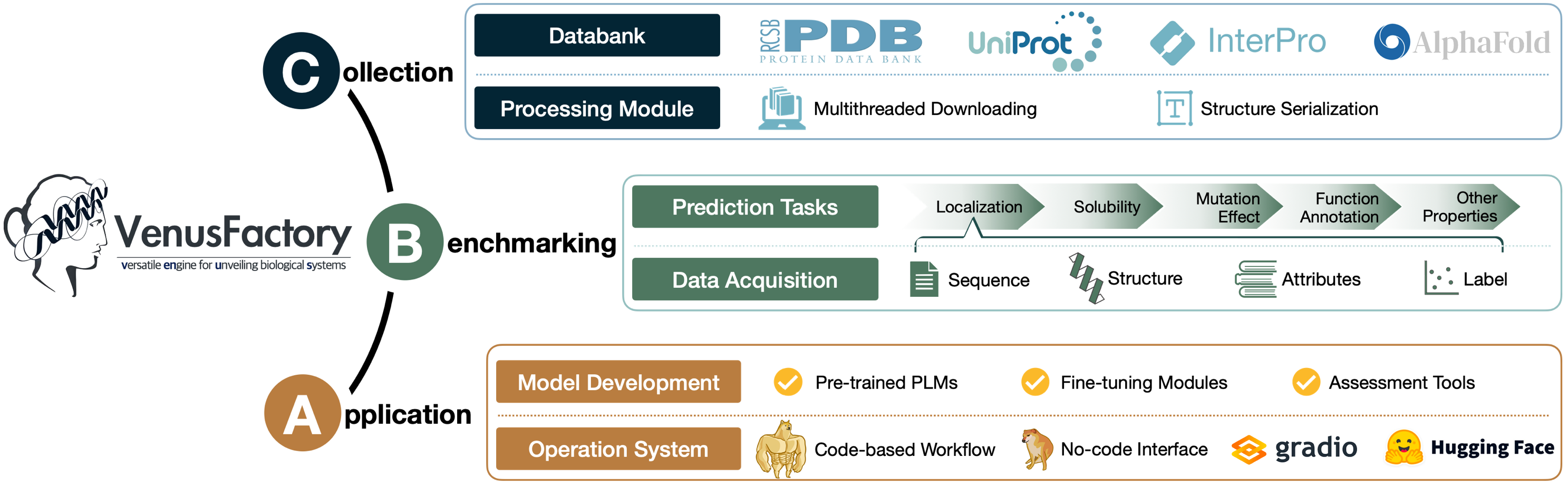}
    \caption{\vf~supports high-throughput raw data download, structure sequencing, a wide range of downstream task datasets, and interface or command-line protein language model fine-tuning and reasoning.}
    \label{fig:architecture}
\end{figure*}

Despite the availability of high-impact models and successful applications in certain scenarios, interdisciplinary collaboration between biologists and computer scientists remains limited. Most algorithm development and validation focus on a few specific benchmarks for particular objectives, while many other datasets and engineering challenges lack readily available tools, even when compatible with existing deep learning methodologies. We attribute this gap to three key complexities: (1) \textbf{Collection}: While some public databanks provide access to protein sequences, structures, and functions, they often lack efficient bulk download options and standardized formatting, which are essential for computer scientists to train PLMs. (2) \textbf{Benchmarking}: AI-driven protein engineering lacks a systematic framework that consolidates benchmarks and baselines. As a result, benchmark datasets from experimental research are underutilized in model development, and state-of-the-art models are rarely integrated into daily research workflows as seamlessly as traditional computational biology tools. (3) \textbf{Application}: Beyond the absence of multifunctional integrated systems, existing PLM solutions often require substantial coding expertise, making them less accessible to non-programmers (\eg, biologists) compared to web-based tools.

To address these challenges, we developed a versatile engine for AI-based protein engineering, namely \vf~(Figure~\ref{fig:architecture}). It integrates a full suite of tools from data acquisition to model training, evaluation, and application. It is designed for users from computer science and biology, regardless of their expertise level in programming. Specifically, \vf~supports \textbf{efficient biological data retrieval} with multithreaded downloading and indexing from major biological databases (\eg, \texttt{RCSB PDB} \citep{burley2019rcsb}, \texttt{UniProt} \citep{uniprot2025uniprot}, \texttt{InterPro} \citep{paysan2023interpro}, and \texttt{AlphaFold DB} \citep{varadi2022alphafolddb}). It also includes implementations for \textbf{comprehensive biological prediction tasks and evaluations} covering solubility, localization, function, and mutation prediction, compiled from 40+ protein-related datasets in a unified format. Moreover, \vf~provides \textbf{effortless PLM implementations} for both pre-trained encoders (\eg, \textsc{ESM2} \citep{lin2023esm2} and \textsc{ProtTrans} \citep{elnaggar2021prottrans}) and downstream task fine-tuning (\eg, \texttt{LoRA} series \citep{hu2022lora,dettmers2023qlora,liu2024dora}, \texttt{Freeze} \& \texttt{Full} fine-tuning, and \texttt{SES-Adapter} \citep{tan2024sesadapter} for protein-related tasks).

To the best of our knowledge, \vf~is the most comprehensive engine for AI-driven protein engineering. It integrates extensive biological data resources, essential processing tools, state-of-the-art PLMs, and fine-tuning modules. It supports both Gradio-based web interface \citep{abid2019gradio} and command-line execution, enabling researchers from both computer science and biology backgrounds to access and utilize its components effortlessly. Built on PyTorch \citep{paszke2019pytorch} and released under the Apache 2.0 license, \vf~ensures broad accessibility and reproducibility, with all datasets and model checkpoints available on Hugging Face.

\section{Data Collection}
The first \texttt{Collection} module enables efficient data retrieval from four major protein databanks. This section outlines its core functionalities and implementation techniques, with additional details provided in Appendix~\ref{app:collection}.

\subsection{Databanks}
\vf~supports data collection from four well-established sources for protein sequences, structures, and functions. (1) \href{https://www.rcsb.org/}{\texttt{RCSB PDB}} contains over $200,000$ experimentally determined atom-level protein 3D structures. (2) \href{https://www.uniprot.org/}{\texttt{UniProt}} provides comprehensive amino acid sequences and functional annotations for over $250$ million proteins curated literature and user submission. (3) \href{https://www.ebi.ac.uk/interpro/}{\texttt{InterPro}} assigns accession numbers and functional descriptions to $\sim41,000$ proteins according to their family, domain, and functional site annotations. (4) \href{https://alphafold.ebi.ac.uk/}{\texttt{AlphaFold DB}} hosts AlphaFold2-predicted 3D structure of proteins from UniProt. It enables structure retrieval by UniProt ID. 

\subsection{Multithreaded Downloading}
The \texttt{Collection} module facilitates multithreaded data downloading by simulating \texttt{HTTP} requests using the  \texttt{requests}, \texttt{fake\_useragent}, and \texttt{concurrent} libraries. Data from \texttt{UniProt} (sequences) and \texttt{AlphaFold DB} (sequences and structures) can be accessed by UniProt IDs, \eg, ``\textit{A0A0C5B5G6}". \texttt{RCSB PDB} is available in multiple formats, including \texttt{.cif}, \texttt{.pdb}, and \texttt{.xml}. All metadata are stored in \texttt{.json} format and indexed by the RCSB ID (\eg, ``\textit{1A00}"). Queryable metadata fields including ``\textit{pubmed\_id}" and ``\textit{assembly\_ids}". For \texttt{InterPro} family data, downloads can be performed using individual InterPro IDs or by parsing family \texttt{.json} files from the website. Retrieved data includes family descriptions (\eg, ``\textit{pfam}" and ``\textit{go\_terms}") as well as detailed protein annotations (\eg, sequence fragments and gene information).

\subsection{Structure Serialization}
Protein structures are crucial for describing protein characteristics, yet structural information alone is often challenging to directly use as input for models like PLMs. \vf~supports conversion tools that encode protein structures into discrete tokens. Three popular serialization methods are considered, including \textsc{DSSP} \citep{kabsch1983dssp}, \textsc{Foldseek} \citep{van2024foldseek}, and the \textsc{ESM3} encoder \citep{Thomas2025esm3}. \textsc{DSSP} converts structures into $3$-class or $8$-class secondary structure representations. \textsc{Foldseek} employs \textsc{VQ-VAE} \citep{Oord2017vqvae} to transform continuous structural data into $20$-dimensional 3Di tokens. The \textsc{ESM3} encoder constructs $4,096$-dimensional integer representations for local subgraphs centered on each amino acid.

\begin{table}[]
\resizebox{\linewidth}{!}{
    \begin{tabular}{@{}ll@{}}
    \toprule
    \multicolumn{2}{c}{\textbf{Essential}} \\ \midrule
    aa\_seq              & Amino acid sequence, \eg, \textit{MASG...} \\
    label & Target label, integer, float, or list, \eg, \textit{0} \\ \midrule
    \multicolumn{2}{c}{\textbf{Optional}} \\ \midrule
    name & Unique Protein or Uniprot ID, \eg, \textit{P05798} \\
    ss3\_seq & 3-class of DSSP sequence, \eg, \textit{CHHHH...} \\
    ss8\_seq & 8-class of DSSP sequence, \eg, \textit{THLEH...} \\
    foldseek\_seq & Foldseek structure sequence, \eg, \textit{CVFLV...} \\
    esm3\_structure\_seq & ESM3 structure sequence, \eg, \textit{{[}85, 3876, ...{]}} \\
    detail or other & Auxiliary information or detailed description \\ \bottomrule
    \end{tabular}
    }
\caption{Benchmark dataset format example.}
\label{tab:data_example}
\end{table}

\section{Task Benchmarking}
Assessing the predictive accuracy of protein representations extracted by PLMs is crucial for both developing new models and guiding biological applications. \vf~integrates over $40$ benchmark datasets from the literature and categorizes them into five major bioengineering tasks to help users gain a comprehensive understanding of common tasks and access relevant datasets. To enhance usability, we have standardized the data formats for all datasets (Table~\ref{tab:data_example}). We introduce the benchmark datasets for the five classes. Further details are provided in Appendix~\ref{app:benchmark}.

\subsection{Localization}
Protein function is closely linked to its cellular compartment or organelle, where specific physiological conditions enable distinct activities. \vf~curates and refines protein localization datasets from \citet{almagro2017deeploc} and \citet{thumuluri2022deeploc2}, including (1) \textbf{DeepLocBinary}: a binary classification of membrane association, (2) \textbf{DeepLocMulti}: a multi-class classification for precise localization, and (3) \textbf{DeepLoc2Multi}: a multi-label, multi-class classification for complex localization scenarios. All three benchmarks include sequence data and AlphaFold2-predicted structures, with additional ESMFold-predicted structures available for \textbf{DeepLocBinary} and \textbf{DeepLocMulti}.

\subsection{Solubility}
Solubility is a prerequisite for proteins to function \textit{in vitro}. However, many proteins, especially those engineered manually, often face solubility challenges. Therefore, it is crucial to predict the solubility of a protein of interest in terms of reducing experimental costs.  \vf~includes three binary classification benchmarks -- \textbf{DeepSol} \citep{Khurana2018deepsol}, \textbf{DeepSoluE} \citep{wang2023deepsolue}, and \textbf{ProtSolM} \citep{tan2024protsolm} -- as well as one regression benchmark, \textbf{eSol} \citep{chen2021esol}. All datasets include protein structures predicted by ESMFold, with \textbf{eSol} additionally providing AlphaFold2-predicted structures.

\subsection{Annotation}
Accurately predicting protein function is essential for understanding enzymatic activity, molecular interactions, and cellular roles in metabolism, signaling, and regulation \citep{zhou2024mlife}. \vf~includes four multi-class, multi-label prediction benchmarks from \citet{su2024saprot}: \textbf{EC}, which uses Enzyme Commission numbers \citep{bairoch2000ec} as function annotation labels; and \textbf{GO-CC}, \textbf{GO-BP}, and \textbf{GO-MF}, which employ Gene Ontology annotations \citep{ashburner2000go}. For all four benchmarks, protein structures are generated using AlphaFold2 and ESMFold.

\begin{table*}[]
\begin{center}
\resizebox{\textwidth}{!}{
    \begin{tabular}{llccccccccccc}
    \toprule
    \multirow{2}{*}{\textbf{Model}} & \multirow{2}{*}{\textbf{Fine-tuning}} & \multicolumn{3}{c}{\textbf{Localization}} & \multicolumn{4}{c}{\textbf{Solubility}} & \multicolumn{4}{c}{\textbf{Annotation}} \\ \cmidrule(l){3-5} \cmidrule(l){6-9} \cmidrule(l){10-13} 
     &  & \multicolumn{1}{c}{\textbf{DL2M}} & \multicolumn{1}{c}{\textbf{DLB}} & \multicolumn{1}{c}{\textbf{DLM}} & \multicolumn{1}{c}{\textbf{DS}} & \multicolumn{1}{c}{\textbf{DSE}} & \multicolumn{1}{c}{\textbf{PSM}} & \multicolumn{1}{c}{\textbf{ES}} & \multicolumn{1}{c}{\textbf{EC}} & \multicolumn{1}{c}{\textbf{BP}} & \multicolumn{1}{c}{\textbf{CC}} & \multicolumn{1}{c}{\textbf{MF}} \\ 
    \midrule 
    \multirow{3}{*}{ESM2-650M} & Freeze & 81.22 & 90.97 & 80.63 & 66.52 & \underline{54.58} & 64.63 & 73.16 & 84.32 & \underline{48.36} & \underline{57.74} & 63.99 \\
     & LoRA & \underline{81.74} & 93.40 & \underline{83.04} & 74.41 & 54.23 & 64.30 & \underline{74.15} & \underline{85.15} & 48.31 & 46.09 & \underline{66.42} \\
     & SES-Adapter & 80.00 & \underline{93.50} & 82.90 & \underline{75.51} & 54.23 & \underline{65.88} & 72.47 & 84.80 & 46.63 & 52.59 & 63.38 \\
     \midrule
    \multirow{3}{*}{Ankh-Large} & Freeze & 79.51 & 90.34 & 80.53 & 64.82 & \underline{\textbf{55.52}} & 64.40 & 71.49 & 85.14 & 45.90 & \underline{54.70} & 61.29 \\
     & LoRA & 76.39 & \underline{\textbf{93.69}} & \underline{83.04} & \underline{74.06} & 55.19 & \underline{66.71} & \underline{\textbf{76.16}} & 75.58 & 28.68 & 38.15 & 48.62 \\
     & SES-Adapter & \underline{81.11} & 92.71 & 82.93 & 73.16 & 55.13 & 66.59 & 69.12 & \underline{86.03} & \underline{47.54} & 49.64 & \underline{64.48} \\
     \midrule
    \multirow{3}{*}{ProtBert} & Freeze & 77.85 & 87.85 & 74.54 & 66.32 & 53.55 & 61.79 & \underline{69.59} & 70.08 & \underline{42.04} & \underline{54.55} & 52.31 \\
     & LoRA & 43.25 & 92.30 & \underline{78.59} & \underline{\textbf{75.81}} & \underline{55.32} & \underline{62.34} & 66.22 & 76.41 & 24.52 & 31.61 & 16.09 \\
     & SES-Adapter & \underline{78.85} & \underline{92.71} & 77.57 & 74.76 & 54.94 & \underline{62.34} & 67.07 & \underline{76.56} & 41.47 & 49.52 & \underline{54.58} \\
     \midrule
    \multirow{3}{*}{ProtT5-XL-U50} & Freeze & 82.50 & 91.78 & 81.18 & 69.22 & \underline{55.13} & 66.08 & \underline{73.22} & 82.57 & 48.84 & \underline{\textbf{59.07}} & 64.39 \\
     & LoRA & 81.94 & \underline{93.11} & 84.06 & 74.86 & 54.03 & 65.17 & 72.77 & \underline{\textbf{87.35}} & 46.40 & 56.55 & \underline{\textbf{67.35}} \\
     & SES-Adapter & \underline{\textbf{82.89}} & 92.71 & \underline{\textbf{85.19}} & \underline{75.26} & 54.94 & \underline{\textbf{67.59}} & 73.11 & 84.56 & \underline{\textbf{49.49}} & 56.86 & 65.11 \\ 
     \bottomrule
    \end{tabular}
    }
\caption{Performance comparison with highlighted best results of \underline{each model} and \textbf{each task}. The detail and evaluation metrics of the dataset can be found in Appendix \ref{app:benchmark}.}
\label{tab:performance1}
\end{center}
\end{table*}

\subsection{Mutation}
Mutating amino acids is a key approach in protein engineering for modifying protein function and properties, such as enzymatic activity, stability, selectivity, and molecular interactions. \vf~includes a total of 19 benchmark datasets with numeric labels, making them suitable for regression tasks. Specifically, we incorporate three enzyme solubility benchmarks from \citet{tan2023peta} (\textbf{PETA\_TEM\_Sol}, \textbf{PETA\_CHS\_Sol}, and \textbf{PETA\_LGK\_Sol}), fluorescence intensity and stability benchmark from \citet{rao2019tape} (\textbf{TAPE\_Fluorescence} and \textbf{TAPE\_Stability}), as well as seven adeno-associated virus fitness benchmarks (\textbf{FLIP\_AAV}) and five nucleotide-binding protein benchmarks (\textbf{FLIP\_GB1}) from \citet{dallago2021flip} with clearly defined splitting rules, such as one-vs-rest training and random sampling.

\subsection{Other Properties}
Beyond the commonly explored tasks and open benchmarks, we have curated five additional datasets that characterize other protein properties. One dataset focuses on stability prediction \textbf{Thermostability} \citep{su2024saprot}. The second \textbf{DeepET\_Topt} \citep{li2022deepet} provides optimal temperature predictions for enzymes. Additionally, we include two binary classification tasks: \textbf{MetalIonBinding} \citep{hu2022revise-plm}, which identifies metal ion-protein binding, and \textbf{SortingSignal} \citep{thumuluri2022deeploc2}, which detects sorting signals involved in protein localization. All datasets incorporate AlphaFold2-predicted structures. Furthermore, \textbf{Thermostability}, \textbf{DeepET\_Topt}, and \textbf{SortingSignal} also include structures by ESMFold.

\section{Model Application}
While many PLMs have been developed, bridging them to biological applications requires applying them to downstream tasks. This involves seamlessly accessing pre-trained PLMs and integrating them with appropriate fine-tuning modules for task-specific training and inference. To facilitate this, \vf~provides a dedicated \texttt{Application} module with specific architectures and optimization strategies to improve performance across diverse tasks.

\subsection{Pre-trained PLMs}
\vf~supports fine-tuning across two primary categories of over $40$ Transformer-based PLMs: Encoder-Only and Encoder-Decoder models. The Encoder-Only category includes both classic and state-of-the-art models, including \textsc{ESM2} (ranging from 8M to 15B parameters) \citep{lin2023esm2}, \textsc{ESM-1b} \citep{rives2021esm1b}, \textsc{ESM-1v} \citep{meier2021esm1v}, \textsc{ProtBert} \citep{elnaggar2021prottrans}, \textsc{IgBert} \citep{kenlay2024igbert}, \textsc{ProSST} \citep{li2024prosst}, \textsc{PETA} \citep{tan2023peta},$40+$  and \textsc{ProPrime} \citep{jiang2024prime}. For Encoder-Decoder architectures, \vf~incorporates models including the \textsc{Ankh} series \citep{elnaggar2023ankh}, \textsc{ProtT5} \citep{elnaggar2021prottrans}, and \textsc{IgT5} \citep{kenlay2024igbert}. Further details can be found in Appendix~\ref{app:models}.

\begin{table*}[]
\begin{center}
\resizebox{\textwidth}{!}{
    \begin{tabular}{llccccccccccc}
    \toprule
    \multirow{2}{*}{\textbf{Model}} & \multirow{2}{*}{\textbf{Fine-tuning}} & \multicolumn{7}{c}{\textbf{Mutation}} & \multicolumn{4}{c}{\textbf{Other}} \\ \cmidrule(l){3-9} \cmidrule(l){10-13} 
     &  & \multicolumn{1}{c}{\textbf{CHS}} & \multicolumn{1}{c}{\textbf{LGK}} & \multicolumn{1}{c}{\textbf{TEM}} & \multicolumn{1}{c}{\textbf{AAV}} & \multicolumn{1}{c}{\textbf{GB1}} & \multicolumn{1}{c}{\textbf{STA}} & \multicolumn{1}{c}{\textbf{FLU}} & \multicolumn{1}{c}{\textbf{SIG}} & \multicolumn{1}{c}{\textbf{MIB}} & \multicolumn{1}{c}{\textbf{DET}} & \multicolumn{1}{c}{\textbf{TMO}} \\ 
     \midrule
    \multirow{3}{*}{ESM2-650M} & Freeze & 26.68 & 27.74 & 13.93 & 70.58 & 71.48 & 68.33 & 45.32 & 88.72 & 67.82 & 67.15 & 68.85 \\
     & LoRA & \underline{35.66} & \underline{30.17} & \underline{30.37} & \underline{93.75} & \underline{93.96} & \underline{78.16} & \underline{50.69} & 90.09 & \underline{73.38} & 60.59 & \underline{\textbf{70.80}} \\
     & SES-Adapter & - & - & - & - & - & - & - & \underline{90.83} & 68.87 & \underline{68.22} & 66.32 \\
     \midrule
    \multirow{3}{*}{Ankh-Large} & Freeze & 32.33 & \underline{41.23} & 20.33 & 69.23 & 76.32 & \underline{67.54} & 52.50 & 84.41 & 75.49 & 64.31 & 66.52 \\
     & LoRA & \underline{37.48} & 36.27 & \underline{20.52} & \underline{93.89} & \underline{94.60} & 62.95 & \underline{68.13} & 87.63 & 74.07 & \underline{64.84} & \underline{69.68} \\
     & SES-Adapter & - & - & - & - & - & - & - & \underline{\textbf{91.35}} & \underline{\textbf{78.35}} & 63.71 & 69.21 \\
     \midrule
    \multirow{3}{*}{ProtBert} & Freeze & 13.49 & \underline{20.50} & \underline{15.51} & 65.96 & 67.26 & 65.35 & \underline{43.73} & 84.83 & 66.77 & 64.83 & 65.58 \\
     & LoRA & \underline{19.22} & 10.56 & 14.09 & \underline{94.05} & \underline{94.41} & \underline{75.11} & 42.85 & 87.22 & \underline{68.42} & 64.82 & \underline{67.05} \\
     & SES-Adapter & - & - & - & - & - & - & - & \underline{90.94} & 67.97 & \underline{64.84} & 66.68 \\
     \midrule
    \multirow{3}{*}{ProtT5-XL-U50} & Freeze & 37.58 & \underline{38.78} & 31.10 & 63.62 & 75.52 & 74.50 & 48.46 & 88.17 & 75.79 & 69.15 & 69.15 \\
     & LoRA & \underline{43.84} & 27.06 & \underline{34.68} & \underline{94.09} & \underline{95.13} & \underline{83.50} & \underline{66.00} & 89.13 & \underline{76.69} & 67.42 & 68.46 \\
     & SES-Adapter & - & - & - & - & - & - & - & \underline{\textbf{91.35}} & 74.14 & \underline{\textbf{70.70}} & \underline{69.71} \\ 
     \bottomrule
    \end{tabular}
    }
\caption{Performance comparison with highlighted best results of \underline{each model} and \textbf{each task}. The detail and evaluation metrics of the dataset can be found in Appendix \ref{app:benchmark}.}
\label{tab:performance2}
\end{center}
\end{table*}

\paragraph{Collate Function}
When training a PLM, protein sequences are typically truncated based on batch size, similar to operations in NLP. However, proteins are complex systems where subtle token replacements can lead to significant functional and structural changes. Additionally, their intrinsic spatial characteristics introduce long-range dependencies between tokens. To address these factors, \vf~supports not only conventional sequence truncation but also a non-truncating approach, which statistically determines an optimal token limit per batch to maintain sequence integrity during training.

\paragraph{Normalization}
We provide multiple normalization methods to enhance training stability and convergence. Supported options include Min-Max normalization, Z-score standardization, Robust normalization, Log transformation, and Quantile normalization.

\subsection{Fine-tuning Modules}
For fine-tuning pre-trained PLMs, \vf~supports two classic approaches: \texttt{freeze} fine-tuning and \texttt{full} fine-tuning, along with various \texttt{LoRA}-based efficient training methods \citep{hu2022lora,dettmers2023qlora,liu2024dora} and a protein-specific \textsc{SES-Adapter} method \citep{tan2024sesadapter} (see Table~\ref{tab:training_methods} for a complete list). Specifically, \texttt{freeze} fine-tuning keeps PLM parameters fixed while updating only the readout layers, whereas \texttt{full} fine-tuning updates the entire model. \texttt{LoRA} and its variants enable parameter-efficient fine-tuning to reduce computational costs, and \textsc{SES-Adapter} employs cross-attention between PLM representations and sequence-structure embeddings (\eg, from \textsc{Foldseek}) to enhance protein-specific fine-tuning.

\paragraph{Classification Head}
\vf~supports three classification heads: a two-layer fully connected network with average pooling, dropout, and GeLU activation; a lightweight head \citep{stark2021lightweight_attn} that combines 1D convolutional feature extraction with attention-weighted pooling for efficient sequence aggregation; and \textsc{Attention1D} \citep{tan2024sesadapter} that employs masked 1D convolution-based attention pooling and a nonlinear projection layer for multi-class classification.

\subsection{Performance Assessment}
\paragraph{Loss Function}
For model training and validation, various loss functions are selected based on the prediction task. \texttt{MSELoss} is used for regression tasks, \texttt{BCEWithLogitsLoss} is applied to multi-class and multi-label tasks, and \texttt{CrossEntropyLoss} is employed for the rest classification tasks.

\paragraph{Evaluation Metrics}
\vf~supports a diverse set of evaluation metrics for robust assessment. For numeric labels, \texttt{Spearman's} $\rho$ and \texttt{MSE} are used to evaluate ranking consistency and quantify prediction differences from the ground truth. For classification tasks, standard metrics such as \texttt{accuracy}, \texttt{precision}, \texttt{recall}, \texttt{F1-score}, \texttt{MCC}, and \texttt{AUROC} are included. Specifically, multi-label classification is assessed using the \texttt{F1-max score}. Further details are in Appendix~\ref{app:metrics}.

\section{Experiments}
We evaluate a range of models across various downstream tasks to demonstrate the practicality of \vf~in integrating diverse models, benchmarks, and fine-tuning strategies. Appendix~\ref{app:benchmark} provides additional information on the selected evaluation datasets, partitioning strategies, and monitored metrics.

\subsection{Experimental Setup}
All fine-tuning methods follow a standardized setup: Each batch is constrained to a maximum of $12,000$ tokens to accommodate long protein sequences, with gradient accumulation set to $8$, effectively yielding a batch size of approximately $200$. The \textsc{AdamW} optimizer \citep{Loshchilov2017adamw} is used with a learning rate of $0.0005$. Training runs for a maximum of $100$ epochs, with early stopping applied if no improvement is observed for $10$ consecutive epochs. To ensure reproducibility, the random seed is set to $3407$. For the \textsc{SES-Adapter} method, input structural sequences are derived from \textsc{Foldseek} and \textsc{DSSP} 8-class representations. All experiments are conducted on a cluster of $20$ RTX 3090 GPUs over two months.

\subsection{Results}
We evaluate different PLMs across multiple tasks using three fine-tuning strategies: \texttt{Freeze}, \texttt{LoRA} (vanilla), and \textsc{SES-adapter} (Tables \ref{tab:performance1}-\ref{tab:performance2}). \textsc{SES-adapter} consistently outperforms other methods, particularly in solubility prediction (\textbf{DSE}, \textbf{PSM}) and mutation effect prediction (\textbf{AAV}, \textbf{GB1}). \texttt{LoRA} demonstrates strong performance in localization tasks and achieves the highest scores for \textbf{DLB}, but exhibits less consistency across solubility and annotation tasks. \texttt{Freeze} generally yields the lowest performance, especially in annotation tasks (\textbf{BP}, \textbf{MF}), but remains competitive in EC classification. 

From a within-model perspective, \textsc{ProtT5-XL-U50} achieves the highest overall performance, particularly excelling in annotation and mutation prediction, while \textsc{Ankh-Large} and \textsc{ESM2-650M} perform comparably but show task-dependent variations. In contrast, \textsc{ProtBert} underperforms in mutation prediction and certain annotation tasks, suggesting potential limitations in capturing functional variations. From a within-fine-tuning perspective, \textsc{SES-adapter} consistently provides the best results across different models, demonstrating its robustness for protein-related tasks. \texttt{LoRA} exhibits strong performance in specific tasks, such as localization, but lacks stability across broader benchmarks. The \texttt{Freeze} method exhibits the largest performance gap across tasks, indicating that \texttt{full} fine-tuning or lightweight adaptation is essential for optimal PLM performance in protein engineering. These results highlight the importance of both model selection and fine-tuning strategies, emphasizing that the optimal configuration should be task-specific to maximize predictive accuracy and generalization.

\section{Related Work}
The use of platforms for LLM fine-tuning and benchmarking has become a widely adopted routine in NLP to accommodate users with diverse domain expertise and programming backgrounds. \textsc{LlamaFactory} \citep{zheng2024llamafactory}, \textsc{Janus} \citep{chen2024janus} integrate multiple efficient fine-tuning methods with a no-code interface, while \textsc{LLaMA-Adapter} \citep{zhang2024llamaadapter}, \textsc{Fast-Chat} \citep{zheng2023fastchat}, and \textsc{LMFlow} \citep{diao2024lmflow} enable lightweight adaptation for instruction-following and multi-modal tasks.

In biology, existing systems primarily focus on protein data integration \citep{szklarczyk2019string,burley2019rcsb,paysan2023interpro,uniprot2025uniprot} and visualization \citep{humphrey1996vmd,delano2002pymol,pettersen2004ucsf_chimera,bobrov2024drugwatch}. For AI-driven protein engineering, only a few platforms offer specialized functionality. \textsc{ProteusAI} \citep{funk_proteusai_2024} streamlines the protein engineering pipeline by establishing an iterative cycle from mutant design to experimental feedback. \textsc{SaprotHub} \citep{su_saprothub_2024}, built upon \textsc{SaProt} \citep{su2024saprot}, provides a Colab-based interface for model training and sharing. In comparison, \vf~is the first platform to support a broader range of PLMs and fine-tuning strategies while also incorporating database scraping and standardized benchmark construction, making it a comprehensive tool for protein-related AI applications.

\section{Conclusion and Discussion}
This work introduces \vf, a versatile engine for unveiling biological systems, offering the most comprehensive resources to date for AI-driven protein engineering. By integrating data collection, benchmarking, and application modules for both pre-trained PLMs and fine-tuning strategies, \vf~enables researchers in computer science and computational biology to efficiently access open-source datasets and develop models for diverse protein-related tasks. Future iterations will expand its capabilities with generative modeling for \textit{de novo} protein design, improved fine-tuning efficiency through advanced adaptation techniques, and broader protein function prediction tasks. We aim to provide a more accessible and powerful tool for researchers at the intersection of AI and biology, fostering innovation and discovery even with minimal computational expertise.

\section*{Acknowledgements}
This work was supported by the grants from the National Science Foundation of China (Grant Number 62302291, 12104295), the Computational Biology Key Program of Shanghai Science and Technology Commission (23JS1400600), Shanghai Jiao Tong University Scientific and Technological Innovation Funds (21X010200843), and Science and Technology Innovation Key R\&D Program of Chongqing (CSTB2022TIAD-STX0017), the Student Innovation Center at Shanghai Jiao Tong University, and Shanghai Artificial Intelligence Laboratory.

\bibliography{1reference}

\begin{thebibliography}{60}
\providecommand{\natexlab}[1]{#1}

\bibitem[{Abid et~al.(2019)Abid, Abdalla, Abid, Khan, Alfozan, and Zou}]{abid2019gradio}
Abubakar Abid, Ali Abdalla, Ali Abid, Dawood Khan, Abdulrahman Alfozan, and James Zou. 2019.
\newblock \href {https://arxiv.org/abs/1906.02569} {{Gradio}: Hassle-free sharing and testing of ml models in the wild}.
\newblock \emph{arXiv:1906.02569}.

\bibitem[{Almagro~Armenteros et~al.(2017)Almagro~Armenteros, S{\o}nderby, S{\o}nderby, Nielsen, and Winther}]{almagro2017deeploc}
Jos{\'e}~Juan Almagro~Armenteros, Casper~Kaae S{\o}nderby, S{\o}ren~Kaae S{\o}nderby, Henrik Nielsen, and Ole Winther. 2017.
\newblock \href {https://academic.oup.com/bioinformatics/article/33/21/3387/3931857} {{DeepLoc}: prediction of protein subcellular localization using deep learning}.
\newblock \emph{Bioinformatics}, 33(21):3387--3395.

\bibitem[{Ashburner et~al.(2000)Ashburner, Ball, Blake, Botstein, Butler, Cherry, Davis, Dolinski, Dwight, Eppig et~al.}]{ashburner2000go}
Michael Ashburner, Catherine~A Ball, Judith~A Blake, David Botstein, Heather Butler, J~Michael Cherry, Allan~P Davis, Kara Dolinski, Selina~S Dwight, Janan~T Eppig, et~al. 2000.
\newblock \href {https://www.nature.com/articles/ng0500_25} {Gene ontology: tool for the unification of biology}.
\newblock \emph{Nature genetics}, 25(1):25--29.

\bibitem[{Bairoch(2000)}]{bairoch2000ec}
Amos Bairoch. 2000.
\newblock \href {https://academic.oup.com/nar/article/28/1/304/2384392} {The {ENZYME} database in 2000}.
\newblock \emph{Nucleic Acids Research}, 28(1):304--305.

\bibitem[{Bobrov et~al.(2024)Bobrov, Saltenis, Sun, Pergola, and He}]{bobrov2024drugwatch}
Artem Bobrov, Domantas Saltenis, Zhaoyue Sun, Gabriele Pergola, and Yulan He. 2024.
\newblock \href {https://doi.org/10.18653/v1/2024.acl-demos.18} {{D}rug{W}atch: A comprehensive multi-source data visualisation platform for drug safety information}.
\newblock In \emph{Proceedings of the 62nd Annual Meeting of the Association for Computational Linguistics (Volume 3: System Demonstrations)}, pages 180--189, Bangkok, Thailand. Association for Computational Linguistics.

\bibitem[{Brown et~al.(2020)Brown, Mann, Ryder, Subbiah, Kaplan, Dhariwal, Neelakantan, Shyam, Sastry, Askell et~al.}]{brown2020gpt3}
Tom Brown, Benjamin Mann, Nick Ryder, Melanie Subbiah, Jared~D Kaplan, Prafulla Dhariwal, Arvind Neelakantan, Pranav Shyam, Girish Sastry, Amanda Askell, et~al. 2020.
\newblock \href {https://proceedings.neurips.cc/paper_files/paper/2020/file/1457c0d6bfcb4967418bfb8ac142f64a-Paper.pdf} {Language models are few-shot learners}.
\newblock \emph{Advances in Neural Information Processing Systems}, 33:1877--1901.

\bibitem[{Burley et~al.(2019)Burley, Berman, Bhikadiya, Bi, Chen, Di~Costanzo, Christie, Dalenberg, Duarte, Dutta et~al.}]{burley2019rcsb}
Stephen~K Burley, Helen~M Berman, Charmi Bhikadiya, Chunxiao Bi, Li~Chen, Luigi Di~Costanzo, Cole Christie, Ken Dalenberg, Jose~M Duarte, Shuchismita Dutta, et~al. 2019.
\newblock \href {https://academic.oup.com/nar/article/47/D1/D464/5144139} {{RCSB Protein Data Bank}: biological macromolecular structures enabling research and education in fundamental biology, biomedicine, biotechnology and energy}.
\newblock \emph{Nucleic Acids Research}, 47(D1):D464--D474.

\bibitem[{Chen et~al.(2021)Chen, Zheng, Zhao, and Yang}]{chen2021esol}
Jianwen Chen, Shuangjia Zheng, Huiying Zhao, and Yuedong Yang. 2021.
\newblock \href {https://jcheminf.biomedcentral.com/articles/10.1186/s13321-021-00488-1} {Structure-aware protein solubility prediction from sequence through graph convolutional network and predicted contact map}.
\newblock \emph{Journal of cheminformatics}, 13:1--10.

\bibitem[{Chen et~al.(2024)Chen, Tang, Zhu, Yan, Jin, Wang, Su, Zhang, Wang, and Tang}]{chen2024janus}
Xiaoyi Chen, Siyuan Tang, Rui Zhu, Shijun Yan, Lei Jin, Zihao Wang, Liya Su, Zhikun Zhang, XiaoFeng Wang, and Haixu Tang. 2024.
\newblock \href {https://dl.acm.org/doi/pdf/10.1145/3658644.3690325} {The {Janus} interface: How fine-tuning in large language models amplifies the privacy risks}.
\newblock In \emph{Proceedings of the 2024 on ACM SIGSAC Conference on Computer and Communications Security}, pages 1285--1299.

\bibitem[{Consortium(2025)}]{uniprot2025uniprot}
UniProt Consortium. 2025.
\newblock \href {https://academic.oup.com/nar/article/53/D1/D609/7902999} {{UniProt}: the universal protein knowledgebase in 2025}.
\newblock \emph{Nucleic Acids Research}, 53(D1):D609--D617.

\bibitem[{Dallago et~al.(2021)Dallago, Mou, Johnston, Wittmann, Bhattacharya, Goldman, Madani, and Yang}]{dallago2021flip}
Christian Dallago, Jody Mou, Kadina~E Johnston, Bruce Wittmann, Nick Bhattacharya, Samuel Goldman, Ali Madani, and Kevin~K Yang. 2021.
\newblock \href {https://openreview.net/forum?id=p2dMLEwL8tF} {{FLIP}: Benchmark tasks in fitness landscape inference for proteins}.
\newblock In \emph{Advance in Neural Information Processing Systems Datasets and Benchmarks Track (Round 2)}.

\bibitem[{DeLano et~al.(2002)}]{delano2002pymol}
Warren~L DeLano et~al. 2002.
\newblock \href {https://citeseerx.ist.psu.edu/document?repid=rep1&type=pdf&doi=ab82608e9a44c17b60d7f908565fba628295dc72#page=44} {{Pymol}: An open-source molecular graphics tool}.
\newblock \emph{CCP4 Newsl. Protein Crystallogr}, 40(1):82--92.

\bibitem[{Dettmers et~al.(2023)Dettmers, Pagnoni, Holtzman, and Zettlemoyer}]{dettmers2023qlora}
Tim Dettmers, Artidoro Pagnoni, Ari Holtzman, and Luke Zettlemoyer. 2023.
\newblock \href {https://dl.acm.org/doi/10.5555/3666122.3666563} {Qlora: Efficient finetuning of quantized llms}.
\newblock \emph{Advances in neural information processing systems}, 36:10088--10115.

\bibitem[{Diao et~al.(2024)Diao, Pan, Dong, Shum, Zhang, Xiong, and Zhang}]{diao2024lmflow}
Shizhe Diao, Rui Pan, Hanze Dong, KaShun Shum, Jipeng Zhang, Wei Xiong, and Tong Zhang. 2024.
\newblock \href {https://doi.org/10.18653/v1/2024.naacl-demo.12} {{LMF}low: An extensible toolkit for finetuning and inference of large foundation models}.
\newblock In \emph{Proceedings of the 2024 Conference of the North American Chapter of the Association for Computational Linguistics: Human Language Technologies (Volume 3: System Demonstrations)}, pages 116--127, Mexico City, Mexico. Association for Computational Linguistics.

\bibitem[{Elnaggar et~al.(2023)Elnaggar, Essam, Salah-Eldin, Moustafa, Elkerdawy, Rochereau, and Rost}]{elnaggar2023ankh}
Ahmed Elnaggar, Hazem Essam, Wafaa Salah-Eldin, Walid Moustafa, Mohamed Elkerdawy, Charlotte Rochereau, and Burkhard Rost. 2023.
\newblock \href {https://arxiv.org/pdf/2301.06568} {Ankh: Optimized protein language model unlocks general-purpose modelling}.
\newblock \emph{arXiv:2301.06568}.

\bibitem[{Elnaggar et~al.(2021)Elnaggar, Heinzinger, Dallago, Rehawi, Wang, Jones, Gibbs, Feher, Angerer, Steinegger et~al.}]{elnaggar2021prottrans}
Ahmed Elnaggar, Michael Heinzinger, Christian Dallago, Ghalia Rehawi, Yu~Wang, Llion Jones, Tom Gibbs, Tamas Feher, Christoph Angerer, Martin Steinegger, et~al. 2021.
\newblock \href {https://ieeexplore.ieee.org/iel7/34/4359286/09477085.pdf} {Prottrans: Toward understanding the language of life through self-supervised learning}.
\newblock \emph{IEEE Transactions on Pattern Analysis and Machine Intelligence}, 44(10):7112--7127.

\bibitem[{Funk et~al.(2024)Funk, Machado, Bradley, Napiorkowska, Gallegos-Dextre, Pashkova, Madsen, Webel, Phaneuf, Jenkins, and Acevedo-Rocha}]{funk_proteusai_2024}
Jonathan Funk, Laura Machado, Samuel~A. Bradley, Marta Napiorkowska, Rodrigo Gallegos-Dextre, Liubov Pashkova, Niklas~G. Madsen, Henry Webel, Patrick~V. Phaneuf, Timothy~P. Jenkins, and Carlos~G. Acevedo-Rocha. 2024.
\newblock \href {https://doi.org/10.1101/2024.10.01.616114} {Proteusai: An open-source and user-friendly platform for machine learning-guided protein design and engineering}.
\newblock In \emph{bioRxiv}.

\bibitem[{Guo et~al.(2025)Guo, Yang, Zhang, Song, Zhang, Xu, Zhu, Ma, Wang, Bi et~al.}]{guo2025deepseek-r1}
Daya Guo, Dejian Yang, Haowei Zhang, Junxiao Song, Ruoyu Zhang, Runxin Xu, Qihao Zhu, Shirong Ma, Peiyi Wang, Xiao Bi, et~al. 2025.
\newblock \href {https://arxiv.org/pdf/2501.12948?} {{Deepseek-r1}: Incentivizing reasoning capability in llms via reinforcement learning}.
\newblock \emph{arXiv:2501.12948}.

\bibitem[{Hayes et~al.(2025)Hayes, Rao, Akin, Sofroniew, Oktay, Lin, Verkuil, Tran, Deaton, Wiggert, Badkundri, Shafkat, Gong, Derry, Molina, Thomas, Khan, Mishra, Kim, Bartie, Nemeth, Hsu, Sercu, Candido, and Rives}]{Thomas2025esm3}
Thomas Hayes, Roshan Rao, Halil Akin, Nicholas~J. Sofroniew, Deniz Oktay, Zeming Lin, Robert Verkuil, Vincent~Q. Tran, Jonathan Deaton, Marius Wiggert, Rohil Badkundri, Irhum Shafkat, Jun Gong, Alexander Derry, Raul~S. Molina, Neil Thomas, Yousuf~A. Khan, Chetan Mishra, Carolyn Kim, Liam~J. Bartie, Matthew Nemeth, Patrick~D. Hsu, Tom Sercu, Salvatore Candido, and Alexander Rives. 2025.
\newblock \href {https://doi.org/10.1126/science.ads0018} {Simulating 500 million years of evolution with a language model}.
\newblock \emph{Science}, 0(0):eads0018.

\bibitem[{Hu et~al.(2022{\natexlab{a}})Hu, yelong shen, Wallis, Allen-Zhu, Li, Wang, Wang, and Chen}]{hu2022lora}
Edward~J Hu, yelong shen, Phillip Wallis, Zeyuan Allen-Zhu, Yuanzhi Li, Shean Wang, Lu~Wang, and Weizhu Chen. 2022{\natexlab{a}}.
\newblock \href {https://openreview.net/forum?id=nZeVKeeFYf9} {Lo{RA}: Low-rank adaptation of large language models}.
\newblock In \emph{International Conference on Learning Representations}.

\bibitem[{Hu et~al.(2022{\natexlab{b}})Hu, Yuan, Yang, Ju, Su, Wang, Yang, and Ding}]{hu2022revise-plm}
Mingyang Hu, Fajie Yuan, Kevin~K Yang, Fusong Ju, Jin Su, Hui Wang, Fei Yang, and Qiuyang Ding. 2022{\natexlab{b}}.
\newblock \href {https://openreview.net/forum?id=U8k0QaBgXS} {Exploring evolution-aware \& -free protein language models as protein function predictors}.
\newblock In \emph{Advances in Neural Information Processing Systems}.

\bibitem[{Humphrey et~al.(1996)Humphrey, Dalke, and Schulten}]{humphrey1996vmd}
William Humphrey, Andrew Dalke, and Klaus Schulten. 1996.
\newblock \href {http://www-s.ks.uiuc.edu/Publications/Papers/PDF/HUMP96/HUMP96.pdf} {{VMD}: visual molecular dynamics}.
\newblock \emph{Journal of molecular graphics}, 14(1):33--38.

\bibitem[{Jiang et~al.(2024)Jiang, Li, Dong, Yu, Sun, Wu, Huang, Kang, Pei, Zhang et~al.}]{jiang2024prime}
Fan Jiang, Mingchen Li, Jiajun Dong, Yuanxi Yu, Xinyu Sun, Banghao Wu, Jin Huang, Liqi Kang, Yufeng Pei, Liang Zhang, et~al. 2024.
\newblock \href {https://www.science.org/doi/full/10.1126/sciadv.adr2641} {A general temperature-guided language model to design proteins of enhanced stability and activity}.
\newblock \emph{Science Advances}, 10(48):eadr2641.

\bibitem[{Kabsch and Sander(1983)}]{kabsch1983dssp}
Wolfgang Kabsch and Christian Sander. 1983.
\newblock \href {https://onlinelibrary.wiley.com/doi/10.1002/bip.360221211} {Dictionary of protein secondary structure: pattern recognition of hydrogen-bonded and geometrical features}.
\newblock \emph{Biopolymers: Original Research on Biomolecules}, 22(12):2577--2637.

\bibitem[{Kang et~al.(2024)Kang, Wu, Zhou, Tan, Kang, Yan, Zong, Li, Liu, and Hong}]{kang2024vhh}
Liqi Kang, Banghao Wu, Bingxin Zhou, Pan Tan, Yun Kang, Yongzhen Yan, Yi~Zong, Shuang Li, Zhuo Liu, and Liang Hong. 2024.
\newblock \href {https://elifesciences.org/reviewed-preprints/102788} {{AI}-enabled alkaline-resistant evolution of protein to apply in mass production}.
\newblock \emph{bioRxiv}, pages 2024--09.

\bibitem[{Kenlay et~al.(2024)Kenlay, Dreyer, Kovaltsuk, Miketa, Pires, and Deane}]{kenlay2024igbert}
Henry Kenlay, Fr{\'e}d{\'e}ric~A Dreyer, Aleksandr Kovaltsuk, Dom Miketa, Douglas Pires, and Charlotte~M Deane. 2024.
\newblock \href {https://journals.plos.org/ploscompbiol/article?id=10.1371/journal.pcbi.1012646} {Large scale paired antibody language models}.
\newblock \emph{PLOS Computational Biology}, 20(12):e1012646.

\bibitem[{Khurana et~al.(2018)Khurana, Rawi, Kunji, Chuang, Bensmail, and Mall}]{Khurana2018deepsol}
Sameer Khurana, Reda Rawi, Khalid Kunji, Gwo-Yu Chuang, Halima Bensmail, and Raghvendra Mall. 2018.
\newblock \href {https://doi.org/10.1093/bioinformatics/bty166} {Deepsol: a deep learning framework for sequence-based protein solubility prediction}.
\newblock \emph{Bioinformatics}, 34(15):2605--2613.

\bibitem[{Li et~al.(2022)Li, Buric, Zrimec, Viknander, Nielsen, Zelezniak, and Engqvist}]{li2022deepet}
Gang Li, Filip Buric, Jan Zrimec, Sandra Viknander, Jens Nielsen, Aleksej Zelezniak, and Martin~KM Engqvist. 2022.
\newblock \href {https://onlinelibrary.wiley.com/doi/pdfdirect/10.1002/pro.4480} {Learning deep representations of enzyme thermal adaptation}.
\newblock \emph{Protein Science}, 31(12):e4480.

\bibitem[{Li et~al.(2024)Li, Tan, Ma, Zhong, Yu, Zhou, Ouyang, Zhou, Tan, and Hong}]{li2024prosst}
Mingchen Li, Yang Tan, Xinzhu Ma, Bozitao Zhong, Huiqun Yu, Ziyi Zhou, Wanli Ouyang, Bingxin Zhou, Pan Tan, and Liang Hong. 2024.
\newblock \href {https://openreview.net/forum?id=4Z7RZixpJQ} {Pro{SST}: Protein language modeling with quantized structure and disentangled attention}.
\newblock In \emph{Advances in Neural Information Processing Systems}.

\bibitem[{Lin et~al.(2023)Lin, Akin, Rao, Hie, Zhu, Lu, Smetanin, Verkuil, Kabeli, Shmueli et~al.}]{lin2023esm2}
Zeming Lin, Halil Akin, Roshan Rao, Brian Hie, Zhongkai Zhu, Wenting Lu, Nikita Smetanin, Robert Verkuil, Ori Kabeli, Yaniv Shmueli, et~al. 2023.
\newblock \href {https://www.science.org/doi/10.1126/science.ade2574} {Evolutionary-scale prediction of atomic-level protein structure with a language model}.
\newblock \emph{Science}, 379(6637):1123--1130.

\bibitem[{Liu et~al.(2022)Liu, Tam, Muqeeth, Mohta, Huang, Bansal, and Raffel}]{liu2022few}
Haokun Liu, Derek Tam, Mohammed Muqeeth, Jay Mohta, Tenghao Huang, Mohit Bansal, and Colin~A Raffel. 2022.
\newblock \href {https://openreview.net/forum?id=rBCvMG-JsPd} {Few-shot parameter-efficient fine-tuning is better and cheaper than in-context learning}.
\newblock \emph{Advances in Neural Information Processing Systems}, 35:1950--1965.

\bibitem[{Liu et~al.(2024)Liu, Wang, Yin, Molchanov, Wang, Cheng, and Chen}]{liu2024dora}
Shih-Yang Liu, Chien-Yi Wang, Hongxu Yin, Pavlo Molchanov, Yu-Chiang~Frank Wang, Kwang-Ting Cheng, and Min-Hung Chen. 2024.
\newblock \href {https://arxiv.org/abs/2402.09353} {Dora: Weight-decomposed low-rank adaptation}.
\newblock In \emph{Forty-first International Conference on Machine Learning}.

\bibitem[{Loshchilov et~al.(2017)Loshchilov, Hutter et~al.}]{Loshchilov2017adamw}
Ilya Loshchilov, Frank Hutter, et~al. 2017.
\newblock \href {https://arxiv.org/pdf/1711.05101v2/1000} {Fixing weight decay regularization in adam}.
\newblock \emph{arXiv:1711.05101}, 5.

\bibitem[{Madani et~al.(2023)Madani, Krause, Greene, Subramanian, Mohr, Holton, Olmos, Xiong, Sun, Socher et~al.}]{madani2023progen}
Ali Madani, Ben Krause, Eric~R Greene, Subu Subramanian, Benjamin~P Mohr, James~M Holton, Jose~Luis Olmos, Caiming Xiong, Zachary~Z Sun, Richard Socher, et~al. 2023.
\newblock \href {https://www.nature.com/articles/s41587-022-01618-2} {Large language models generate functional protein sequences across diverse families}.
\newblock \emph{Nature Biotechnology}, 41(8):1099--1106.

\bibitem[{Meier et~al.(2021)Meier, Rao, Verkuil, Liu, Sercu, and Rives}]{meier2021esm1v}
Joshua Meier, Roshan Rao, Robert Verkuil, Jason Liu, Tom Sercu, and Alex Rives. 2021.
\newblock \href {https://proceedings.neurips.cc/paper/2021/hash/f51338d736f95dd42427296047067694-Abstract.html} {Language models enable zero-shot prediction of the effects of mutations on protein function}.
\newblock \emph{Advances in Neural Information Processing Systems}, 34:29287--29303.

\bibitem[{Pan(2023)}]{pan2023llm_chem}
Jie Pan. 2023.
\newblock \href {https://www.nature.com/articles/s43588-023-00399-1} {Large language model for molecular chemistry}.
\newblock \emph{Nature Computational Science}, 3(1):5--5.

\bibitem[{Paszke et~al.(2019)Paszke, Gross, Massa, Lerer, Bradbury, Chanan, Killeen, Lin, Gimelshein, Antiga et~al.}]{paszke2019pytorch}
Adam Paszke, Sam Gross, Francisco Massa, Adam Lerer, James Bradbury, Gregory Chanan, Trevor Killeen, Zeming Lin, Natalia Gimelshein, Luca Antiga, et~al. 2019.
\newblock \href {https://proceedings.neurips.cc/paper_files/paper/2019/file/bdbca288fee7f92f2bfa9f7012727740-Paper.pdf} {Pytorch: An imperative style, high-performance deep learning library}.
\newblock \emph{Advances in neural information processing systems}, 32.

\bibitem[{Paysan-Lafosse et~al.(2023)Paysan-Lafosse, Blum, Chuguransky, Grego, Pinto, Salazar, Bileschi, Bork, Bridge, Colwell et~al.}]{paysan2023interpro}
Typhaine Paysan-Lafosse, Matthias Blum, Sara Chuguransky, Tiago Grego, Beatriz~L{\'a}zaro Pinto, Gustavo~A Salazar, Maxwell~L Bileschi, Peer Bork, Alan Bridge, Lucy Colwell, et~al. 2023.
\newblock \href {https://academic.oup.com/nar/article/51/D1/D418/6814474} {{InterPro} in 2022}.
\newblock \emph{Nucleic Acids Research}, 51(D1):D418--D427.

\bibitem[{Pettersen et~al.(2004)Pettersen, Goddard, Huang, Couch, Greenblatt, Meng, and Ferrin}]{pettersen2004ucsf_chimera}
Eric~F Pettersen, Thomas~D Goddard, Conrad~C Huang, Gregory~S Couch, Daniel~M Greenblatt, Elaine~C Meng, and Thomas~E Ferrin. 2004.
\newblock \href {https://onlinelibrary.wiley.com/doi/full/10.1002/jcc.20084?casa_token=WTOmT7Fee68AAAAA:6mQLZ2pd9mzP06-y9D20zv_eT8DXRdVekxHCV_mJ7w96cqdvV3ar44QXz6N5-SN913ugUIQ0_4HnweA} {{UCSF Chimera}—a visualization system for exploratory research and analysis}.
\newblock \emph{Journal of computational chemistry}, 25(13):1605--1612.

\bibitem[{Rao et~al.(2019)Rao, Bhattacharya, Thomas, Duan, Chen, Canny, Abbeel, and Song}]{rao2019tape}
Roshan Rao, Nicholas Bhattacharya, Neil Thomas, Yan Duan, Peter Chen, John Canny, Pieter Abbeel, and Yun Song. 2019.
\newblock \href {https://proceedings.neurips.cc/paper_files/paper/2019/hash/37f65c068b7723cd7809ee2d31d7861c-Abstract.html} {Evaluating protein transfer learning with tape}.
\newblock \emph{Advances in Neural Information Processing Systems}, 32.

\bibitem[{Rives et~al.(2021)Rives, Meier, Sercu, Goyal, Lin, Liu, Guo, Ott, Zitnick, Ma et~al.}]{rives2021esm1b}
Alexander Rives, Joshua Meier, Tom Sercu, Siddharth Goyal, Zeming Lin, Jason Liu, Demi Guo, Myle Ott, C~Lawrence Zitnick, Jerry Ma, et~al. 2021.
\newblock \href {https://www.pnas.org/doi/10.1073/pnas.2016239118} {Biological structure and function emerge from scaling unsupervised learning to 250 million protein sequences}.
\newblock \emph{Proceedings of the National Academy of Sciences}, 118(15):e2016239118.

\bibitem[{St{\"a}rk et~al.(2021)St{\"a}rk, Dallago, Heinzinger, and Rost}]{stark2021lightweight_attn}
Hannes St{\"a}rk, Christian Dallago, Michael Heinzinger, and Burkhard Rost. 2021.
\newblock \href {https://academic.oup.com/bioinformaticsadvances/article/1/1/vbab035/6432029} {Light attention predicts protein location from the language of life}.
\newblock \emph{Bioinformatics Advances}, 1(1):vbab035.

\bibitem[{Su et~al.(2024{\natexlab{a}})Su, Han, Zhou, Shan, Zhou, and Yuan}]{su2024saprot}
Jin Su, Chenchen Han, Yuyang Zhou, Junjie Shan, Xibin Zhou, and Fajie Yuan. 2024{\natexlab{a}}.
\newblock \href {https://openreview.net/forum?id=6MRm3G4NiU} {{SaProt}: Protein language modeling with structure-aware vocabulary}.
\newblock In \emph{The Twelfth International Conference on Learning Representations}.

\bibitem[{Su et~al.(2024{\natexlab{b}})Su, Li, Han, Zhou, He, Shan, Zhou, Chang, Jiang, Ma, {The OPMC}, Steinegger, Ovchinnikov, and Yuan}]{su_saprothub_2024}
Jin Su, Zhikai Li, Chenchen Han, Yuyang Zhou, Yan He, Junjie Shan, Xibin Zhou, Xing Chang, Shiyu Jiang, Dacheng Ma, {The OPMC}, Martin Steinegger, Sergey Ovchinnikov, and Fajie Yuan. 2024{\natexlab{b}}.
\newblock \href {https://www.biorxiv.org/content/10.1101/2024.05.24.595648v1} {{SaprotHub}: Making protein modeling accessible to all biologists}.
\newblock In \emph{bioRxiv}.

\bibitem[{Szklarczyk et~al.(2019)Szklarczyk, Gable, Lyon, Junge, Wyder, Huerta-Cepas, Simonovic, Doncheva, Morris, Bork et~al.}]{szklarczyk2019string}
Damian Szklarczyk, Annika~L Gable, David Lyon, Alexander Junge, Stefan Wyder, Jaime Huerta-Cepas, Milan Simonovic, Nadezhda~T Doncheva, John~H Morris, Peer Bork, et~al. 2019.
\newblock \href
  {https://watermark.silverchair.com/gky1131.pdf?token=AQECAHi208BE49Ooan9kkhW_Ercy7Dm3ZL_9Cf3qfKAc485ysgAAA00wggNJBgkqhkiG9w0BBwagggM6MIIDNgIBADCCAy8GCSqGSIb3DQEHATAeBglghkgBZQMEAS4wEQQM7oyksXbeID_HXuMnAgEQgIIDADkyWQNCjg-cruJ3yGiuGIFe1Y7pgWiudaxwnJC0-ea2z41T-TvZXoYY7kkJ74qxIor7zbw_Dii9lBGqO7-GJQqudP_vIm0fIkNtaD-8jULiFvFS4pmcd4KMyc-K8y0ssM55NK-MUjQMqg2GtdCsBLnUSg9cUA98q2qkOxi8cM8h9lO3_9P0sGglcC2yE2GRAL9Rz58X6e-8SQQn8CNCpV76-eSQxslgtCtNraaHiE3e50eMij20H7EVQQbaGFH2iUpEu6J8YPZ9G2r2EQ5_rp6GL1Ek7-Uf1VIiJ_98kIVqFaN4ZAJaQXJaXZpwSvRNvhHYeobNIaZX9EpgiSL1JJsIVO6osPKoLrkgKUzuW3eTyIbEV8wVkof24O-N-b7uDrQ0cqzhPFdS4DsFoDKuj0C-7AYylOekPl3swkSFqS6eSv4ByyCJk1aLoS8yQIrWIWTNP8rTNXWZWtpf4q0g5tEMhC_Xv03KKGt7QXLdNMVwcm6SK-NELDUXmxwlJipSCRDZt-xo3-1tcvWARX3w8YgLB70wkXQnR30d5d0ul_K-0MIrmb4r9C_LmW8Y1w4Txqf5wr5CjbTe11DSLtfcuSrX1BipLLiAeHQZzkijixPvN1CKugCm0K0i11yThFEg3nNns_91O3k2Kk5pF39ikyDFO6F9fBsdNCYxsW_fscbtr3iZ9kx-lTGGizHE_rxiFcKpKv1-trLmx9a9yJUEoV7dVuDwgveh1_F3_4VWWEeeiSOkqkUJOR4mmbujcTOoaUCGStiGRbs9XuyvfibjB0fM6or8-5HwNmTWARzDU9DDve6o9O8rCeQZvaW2H-6ivrn45qZxd0USLCbWtbQcCz8lY21ZN_lBjEPOOQogUssWPJU67UlxM61-JskvBgQpxYbUw5W4_5sjOjYDlPq8UY7z_4rxY5dTxauHn2jxpq4oPrGmo5H0AWvT43pzBuW5raDfFX5QvwARU4tHVfJNlYcKkuHfRqpgly94KFYloCaJipQFqpk_TOXAa-bRrQs93A}
  {String v11: protein--protein association networks with increased coverage, supporting functional discovery in genome-wide experimental datasets}.
\newblock \emph{Nucleic acids research}, 47(D1):D607--D613.

\bibitem[{Tan et~al.(2024{\natexlab{a}})Tan, Li, Zhou, Zhong, Zheng, Tan, Zhou, Yu, Fan, and Hong}]{tan2024sesadapter}
Yang Tan, Mingchen Li, Bingxin Zhou, Bozitao Zhong, Lirong Zheng, Pan Tan, Ziyi Zhou, Huiqun Yu, Guisheng Fan, and Liang Hong. 2024{\natexlab{a}}.
\newblock \href {https://pubs.acs.org/doi/10.1021/acs.jcim.4c00689} {Simple, efficient, and scalable structure-aware adapter boosts protein language models}.
\newblock \emph{Journal of Chemical Information and Modeling}.

\bibitem[{Tan et~al.(2024{\natexlab{b}})Tan, Li, Zhou, Tan, Yu, Fan, and Hong}]{tan2023peta}
Yang Tan, Mingchen Li, Ziyi Zhou, Pan Tan, Huiqun Yu, Guisheng Fan, and Liang Hong. 2024{\natexlab{b}}.
\newblock \href {https://jcheminf.biomedcentral.com/articles/10.1186/s13321-024-00884-3} {{PETA}: evaluating the impact of protein transfer learning with sub-word tokenization on downstream applications}.
\newblock \emph{Journal of Cheminformatics}, 16(1):92.

\bibitem[{Tan et~al.(2024{\natexlab{c}})Tan, Zheng, Hong, and Zhou}]{tan2024protsolm}
Yang Tan, Jia Zheng, Liang Hong, and Bingxin Zhou. 2024{\natexlab{c}}.
\newblock \href {https://ieeexplore.ieee.org/document/10822310/} {{ProtSolM}: Protein solubility prediction with multi-modal features}.
\newblock In \emph{2024 IEEE International Conference on Bioinformatics and Biomedicine (BIBM)}, pages 223--232. IEEE.

\bibitem[{Tan et~al.(2023)Tan, Zhou, Zheng, Fan, and Hong}]{tan2023protssn}
Yang Tan, Bingxin Zhou, Lirong Zheng, Guisheng Fan, and Liang Hong. 2023.
\newblock \href {https://www.biorxiv.org/content/10.1101/2023.12.01.569522v1} {Semantical and topological protein encoding toward enhanced bioactivity and thermostability}.
\newblock \emph{bioRxiv}, pages 2023--12.

\bibitem[{Thumuluri et~al.(2022)Thumuluri, Almagro~Armenteros, Johansen, Nielsen, and Winther}]{thumuluri2022deeploc2}
Vineet Thumuluri, Jos{\'e}~Juan Almagro~Armenteros, Alexander~Rosenberg Johansen, Henrik Nielsen, and Ole Winther. 2022.
\newblock \href {https://academic.oup.com/nar/article/50/W1/W228/6576357} {{DeepLoc 2.0}: multi-label subcellular localization prediction using protein language models}.
\newblock \emph{Nucleic Acids Research}, 50(W1):W228--W234.

\bibitem[{van~den Oord et~al.(2017)van~den Oord, Vinyals, and kavukcuoglu}]{Oord2017vqvae}
Aaron van~den Oord, Oriol Vinyals, and koray kavukcuoglu. 2017.
\newblock \href {https://proceedings.neurips.cc/paper_files/paper/2017/file/7a98af17e63a0ac09ce2e96d03992fbc-Paper.pdf} {Neural discrete representation learning}.
\newblock In \emph{Advances in Neural Information Processing Systems}, volume~30. Curran Associates, Inc.

\bibitem[{Van~Kempen et~al.(2024)Van~Kempen, Kim, Tumescheit, Mirdita, Lee, Gilchrist, S{\"o}ding, and Steinegger}]{van2024foldseek}
Michel Van~Kempen, Stephanie~S Kim, Charlotte Tumescheit, Milot Mirdita, Jeongjae Lee, Cameron~LM Gilchrist, Johannes S{\"o}ding, and Martin Steinegger. 2024.
\newblock \href {https://www.nature.com/articles/s41587-023-01773-0} {Fast and accurate protein structure search with {Foldseek}}.
\newblock \emph{Nature Biotechnology}, 42(2):243--246.

\bibitem[{Varadi et~al.(2022)Varadi, Anyango, Deshpande, Nair, Natassia, Yordanova, Yuan, Stroe, Wood, Laydon et~al.}]{varadi2022alphafolddb}
Mihaly Varadi, Stephen Anyango, Mandar Deshpande, Sreenath Nair, Cindy Natassia, Galabina Yordanova, David Yuan, Oana Stroe, Gemma Wood, Agata Laydon, et~al. 2022.
\newblock \href {https://academic.oup.com/nar/article/50/D1/D439/6430488} {Alphafold protein structure database: massively expanding the structural coverage of protein-sequence space with high-accuracy models}.
\newblock \emph{Nucleic Acids Research}, 50(D1):D439--D444.

\bibitem[{Wang and Zou(2023)}]{wang2023deepsolue}
Chao Wang and Quan Zou. 2023.
\newblock \href {https://bmcbiol.biomedcentral.com/articles/10.1186/s12915-023-01510-8} {Prediction of protein solubility based on sequence physicochemical patterns and distributed representation information with deepsolue}.
\newblock \emph{BMC biology}, 21(1):12.

\bibitem[{Zhang et~al.(2023)Zhang, Chen, Bukharin, Karampatziakis, He, Cheng, Chen, and Zhao}]{zhang2023adalora}
Qingru Zhang, Minshuo Chen, Alexander Bukharin, Nikos Karampatziakis, Pengcheng He, Yu~Cheng, Weizhu Chen, and Tuo Zhao. 2023.
\newblock \href {https://arxiv.org/abs/2303.10512} {Adalora: Adaptive budget allocation for parameter-efficient fine-tuning}.
\newblock \emph{arXiv preprint arXiv:2303.10512}.

\bibitem[{Zhang et~al.(2024)Zhang, Han, Liu, Zhou, Lu, Qiao, Li, and Gao}]{zhang2024llamaadapter}
Renrui Zhang, Jiaming Han, Chris Liu, Aojun Zhou, Pan Lu, Yu~Qiao, Hongsheng Li, and Peng Gao. 2024.
\newblock \href {https://openreview.net/forum?id=d4UiXAHN2W} {{LL}a{MA}-adapter: Efficient fine-tuning of large language models with zero-initialized attention}.
\newblock In \emph{International Conference on Learning Representations}.

\bibitem[{Zheng et~al.(2023)Zheng, Chiang, Sheng, Zhuang, Wu, Zhuang, Lin, Li, Li, Xing, Zhang, Gonzalez, and Stoica}]{zheng2023fastchat}
Lianmin Zheng, Wei-Lin Chiang, Ying Sheng, Siyuan Zhuang, Zhanghao Wu, Yonghao Zhuang, Zi~Lin, Zhuohan Li, Dacheng Li, Eric Xing, Hao Zhang, Joseph~E. Gonzalez, and Ion Stoica. 2023.
\newblock \href {https://openreview.net/forum?id=uccHPGDlao} {Judging {LLM}-as-a-judge with {MT}-bench and chatbot arena}.
\newblock In \emph{Advance in Neural Information Processing Systems Datasets and Benchmarks Track}.

\bibitem[{Zheng et~al.(2024)Zheng, Zhang, Zhang, Ye, and Luo}]{zheng2024llamafactory}
Yaowei Zheng, Richong Zhang, Junhao Zhang, Yanhan Ye, and Zheyan Luo. 2024.
\newblock \href {https://doi.org/10.18653/v1/2024.acl-demos.38} {{L}lama{F}actory: Unified efficient fine-tuning of 100+ language models}.
\newblock In \emph{Proceedings of the 62nd Annual Meeting of the Association for Computational Linguistics (Volume 3: System Demonstrations)}, pages 400--410, Bangkok, Thailand. Association for Computational Linguistics.

\bibitem[{Zhou et~al.(2024{\natexlab{a}})Zhou, Tan, Hu, Zheng, Zhong, and Hong}]{zhou2024mlife}
Bingxin Zhou, Yang Tan, Yutong Hu, Lirong Zheng, Bozitao Zhong, and Liang Hong. 2024{\natexlab{a}}.
\newblock \href {https://onlinelibrary.wiley.com/doi/full/10.1002/mlf2.12157} {Protein engineering in the deep learning era}.
\newblock \emph{mLife}, 3(4):477--491.

\bibitem[{Zhou et~al.(2024{\natexlab{b}})Zhou, Zheng, Wu, Yi, Zhong, Tan, Liu, Li{\`o}, and Hong}]{zhou2024cpdiffusion}
Bingxin Zhou, Lirong Zheng, Banghao Wu, Kai Yi, Bozitao Zhong, Yang Tan, Qian Liu, Pietro Li{\`o}, and Liang Hong. 2024{\natexlab{b}}.
\newblock \href {https://www.nature.com/articles/s41421-024-00728-2} {A conditional protein diffusion model generates artificial programmable endonuclease sequences with enhanced activity}.
\newblock \emph{Cell Discovery}, 10(1):95.

\end{thebibliography}

\clearpage
\appendix

\section{Models}\label{app:models}
\begin{table*}[ht]
\begin{center}
\resizebox{\textwidth}{!}{
    \begin{tabular}{lccll}
    \toprule
    \textbf{Model} & \textbf{\# Params.} & \textbf{Num.} & \textbf{Type} & \textbf{Implement} \\
    \midrule
    ESM2 \citep{lin2023esm2} & 8M-15B & 6 & Encoder & \href{https://huggingface.co/facebook/esm2_t33_650M_UR50D}{\texttt{facebook/esm2\_t33\_650M\_UR50D}}\\
    ESM-1b \citep{rives2021esm1b}
    & 650M & 1 & Encoder &\href{https://huggingface.co/facebook/esm1b_t33_650M_UR50S}{\texttt{facebook/esm1b\_t33\_650M\_UR50S}}\\
    ESM-1v \citep{meier2021esm1v} & 650M & 5 & Encoder  & \href{https://hf.co/facebook/esm1v_t33_650M_UR90S_1}{\texttt{facebook/esm1v\_t33\_650M\_UR90S\_1}}\\
    ProtBert-Uniref100 \citep{elnaggar2021prottrans} & 420M & 1 & Encoder  & \href{https://huggingface.co/Rostlab/prot_bert}{\texttt{Rostlab/prot\_bert\_Uniref100}}\\
    ProtBert-BFD100 \citep{elnaggar2021prottrans} & 420M & 1 & Encoder & \href{https://huggingface.co/Rostlab/prot_bert_bfd}{\texttt{Rostlab/prot\_bert\_bfd}}\\
    IgBert \citep{kenlay2024igbert} & 420M & 1 & Encoder & \href{https://huggingface.co/Exscientia/IgBert}{\texttt{Exscientia/IgBert}}\\
    IgBert\_unpaired \citep{kenlay2024igbert} & 420M & 1 & Encoder & \href{https://huggingface.co/Exscientia/IgBert_unpaired}{\texttt{Exscientia/IgBert\_unpaired}}\\
    ProtT5-Uniref50 \citep{elnaggar2021prottrans} & 3B/11B & 2 & Encoder-Decoder  & \href{https://huggingface.co/Rostlab/prot_t5_xl_uniref50}{\texttt{Rostlab/prot\_t5\_xl\_uniref50}}\\
    ProtT5-BFD100 \citep{elnaggar2021prottrans} & 3B/11B & 2 & Encoder-Decoder & \href{https://huggingface.co/Rostlab/prot_t5_xl_bfd}{\texttt{Rostlab/prot\_t5\_xl\_bfd}}\\
    Ankh \citep{elnaggar2023ankh} & 450M/1.2B & 2 & Encoder-Decoder &\href{https://huggingface.co/Rostlab/prot_t5_xl_uniref50}{\texttt{ElnaggarLab/ankh-base}}\\
    ProSST \citep{li2024prosst} & 110M & 7 & Encoder  & \href{https://huggingface.co/AI4Protein/ProSST-2048}{\texttt{AI4Protein/ProSST-2048}}\\
    ProPrime \citep{jiang2024prime} & 690M & 1 & Encoder &\href{https://huggingface.co/AI4Protein/Prime_690M}{\texttt{AI4Protein/Prime\_690M}}\\
    PETA \citep{tan2023peta} & 80M & 15 & Encoder & \href{https://huggingface.co/AI4Protein/deep_base}{\texttt{AI4Protein/deep\_base}} \\
    \bottomrule
    \end{tabular}
}
\caption{Detail of PLMs in terms of parameters, architecture, and implementation sources.}\label{tab:models}
\end{center}
\end{table*}

Table \ref{tab:models} presents an overview of various PLMs used in computational biology and protein engineering.

\section{Training Methods}\label{app:training_methods}
\subsection{Supported Methods}
Table \ref{tab:training_methods} provides an overview of fine-tuning methods used for PLMs, categorized by their adaptation approach. 
\begin{table}[]
\centering
\resizebox{\linewidth}{!}{
    \begin{tabular}{@{}ll@{}}
    \toprule
    \textbf{Fine-tunning Method} & \textbf{Type} \\
    \midrule
    Freeze       & Sequence \\
    Full & Sequence \\
    LoRA \citep{hu2022lora}         & Sequence \\
    DoRA \citep{liu2024dora}        & Sequence \\
    AdaLoRA \citep{zhang2023adalora}      & Sequence \\
    IA3 \citep{liu2022few}         & Sequence \\
    QLoRA \citep{dettmers2023qlora}       & Sequence \\
    SES-Adapter \citep{tan2024sesadapter}  & Sequence \& Structure \\
    \bottomrule
    \end{tabular}
}
\caption{Supported fine-tuning methods with data modality compatibility.}\label{tab:training_methods}
\end{table}

\subsection{Training Parameters}
Tbale \ref{tab:finetune-params} compares the number of trainable parameters and their relative proportion in different PLMs when applying various fine-tuning methods.
\begin{table}[]
\centering
\resizebox{\linewidth}{!}{
    \begin{tabular}{@{}llcc@{}}
    \toprule
    \textbf{Model} & \textbf{Fine-tuning} & \textbf{Params. (M)} & \textbf{Ratio (\%)} \\
    \midrule
    \multirow{3}{*}{ESM2-650M}
     & Freeze & 1.66 & 0.25 \\
     & LoRA          & 3.67 & 0.56 \\
     & SES-Adapter   & 14.86 & 2.23 \\
    \midrule
    \multirow{3}{*}{Ankh-Large}
     & Freeze & 2.38 & 0.21 \\
     & LoRA          & 5.31 & 0.46 \\
     & SES-Adapter   & 21.71 & 1.85 \\
    \midrule
    \multirow{3}{*}{ProtBert}
     & Freeze & 1.06 & 0.25 \\
     & LoRA          & 2.53 & 0.60 \\
     & SES-Adapter   & 9.52 & 2.22 \\
    \midrule
    \multirow{3}{*}{ProtT5-XL-U50}
     & Freeze & 1.05 & 0.09 \\
     & LoRA          & 4.00 & 0.33 \\
     & SES-Adapter   & 9.71 & 0.80 \\
    \bottomrule
    \end{tabular}
}
\caption{The trainable parameters of different models using different fine-tuning methods and their proportion in the total model.}\label{tab:finetune-params}
\end{table}

\begin{table*}[ht]
\begin{center}
\resizebox{\textwidth}{!}{
    \begin{tabular}{@{}lllccccl@{}}
    \toprule
    \textbf{Dataset} & \textbf{AF2\_pLDDT} & \textbf{EF\_pLDDT} & \textbf{Train} & \textbf{Valid} & \textbf{Test} & \textbf{Metrics} & \textbf{Implement} \\ \midrule
    \multicolumn{8}{c}{\textbf{Localization}} \\
    \midrule
    DeepLoc2Multi (DL2M) & $77.46_{(12.51)}$ & - & $21,948$ & $2,744$ & $2,744$ & accuracy & \href{https://huggingface.co/datasets/tyang816/DeepLoc2Multi}{\texttt{tyang816/DeepLoc2Multi}} \\
    DeepLocBinary (DLB) & $79.57_{(12.06)}$ & $77.10_{(14.62)}$ & $5,735$ & $1,009$ & $1,728$ & accuracy & \href{https://huggingface.co/datasets/tyang816/DeepLocBinary}{\texttt{tyang816/DeepLocBinary}} \\
    DeepLocMulti (DLM) & $77.34_{(12.77)}$ & $74.88_{(15.23)}$ & $9,324$ & $1,658$ & $2,742$ & accuracy & \href{https://huggingface.co/datasets/tyang816/DeepLocMulti}{\texttt{tyang816/DeepLocMulti}} \\
    \midrule
    \multicolumn{8}{c}{\textbf{Solubility}} \\
    \midrule
    DeepSol (DS) & - & $79.59_{13.36}$ & $62,478$ & $6,942$ & $2,001$ & accuracy & \href{https://huggingface.co/datasets/tyang816/DeepSol}{\texttt{tyang816/DeepSol}} \\
    DeepSoluE (DSE) & - & $80.68_{(12.79)}$ & $10,290$ & $1,143$ & $3,100$ & accuracy & \href{https://huggingface.co/datasets/tyang816/DeepSoluE}{\texttt{tyang816/DeepSoluE}} \\
    ProtSolM (PSM) & - & $73.80_{(15.51)}$ & $57,725$ & $3,210$ & $3,208$ & accuracy & \href{https://huggingface.co/datasets/tyang816/ProtSolM}{\texttt{tyang816/ProtSolM}} \\
    eSOL (ES) & $90.79_{(7.07)}$ & $83.45_{(10.39)}$ & $2,481$ & $310$ & $310$ & Spearman's $\rho$ & \href{https://huggingface.co/datasets/tyang816/eSOL}{\texttt{tyang816/eSOL}} \\
    \midrule
    \multicolumn{8}{c}{\textbf{Annoation}} \\
    \midrule
    EC & $92.78_{(6.42)}$ & $85.08_{(8.48)}$ & $13,090$ & $1,465$ & $1,604$ & f1\_max & \href{https://huggingface.co/datasets/tyang816/EC}{\texttt{tyang816/EC}} \\
    GO\_MF (MF) & $91.77_{(6.68)}$ & $82.84_{(9.68)}$ & $22,081$ & $2,432$ & $3,350$ & f1\_max & \href{https://huggingface.co/datasets/tyang816/GO_MF}{\texttt{tyang816/GO\_MF}} \\
    GO\_BP (BP) & $91.35_{(7.06)}$ & $82.00_{(10.65)}$ & $20,947$ & $2,334$ & $3,350$ & f1\_max & \href{https://huggingface.co/datasets/tyang816/GO_BP}{\texttt{tyang816/GO\_BP}} \\
    GO\_CC (CC) & $90.07_{(8.05)}$ & $79.57_{(11.61)}$ & $9,552$ & $1,092$ & $3,350$ & f1\_max & \href{https://huggingface.co/datasets/tyang816/GO_CC}{\texttt{tyang816/GO\_CC}} \\
    \midrule
    \multicolumn{8}{c}{\textbf{Mutation}} \\
    \midrule
    PETA\_CHS\_Sol (CHS) & - & - & $3,872$ & $484$ & $484$ & Spearman's $\rho$ & \href{https://huggingface.co/datasets/tyang816/PETA_CHS_Sol}{\texttt{tyang816/PETA\_CHS\_Sol}} \\
    PETA\_LGK\_Sol (LGK) & - & - & $15,308$ & $1,914$ & $1,914$ & Spearman's $\rho$ & \href{https://huggingface.co/datasets/tyang816/PETA_LGK_Sol}{\texttt{tyang816/PETA\_LGK\_Sol}} \\
    PETA\_TEM\_Sol (TEM) & - & - & $6,445$ & $808$ & $808$ & Spearman's $\rho$ & \href{https://huggingface.co/datasets/tyang816/PETA_TEM_Sol}{\texttt{tyang816/PETA\_TEM\_Sol}} \\
    FLIP\_AAV\_sampled (AAV) & - & - & $66,066$ & $16,517$ & $16,517$ & Spearman's $\rho$ & \href{https://huggingface.co/datasets/tyang816/FLIP_AAV_sampled}{\texttt{tyang816/FLIP\_AAV\_sampled}} \\
    FLIP\_GB1\_sampled (GB1) & - & - & $6,988$ & $1,745$ & $1,745$ & Spearman's $\rho$ & \href{https://huggingface.co/datasets/tyang816/FLIP_GB1_sampled}{\texttt{tyang816/FLIP\_GB1\_sampled}} \\
    TAPE\_Stablity (STA) & - & - & $53,614$ & $2,512$ & $12,851$ & Spearman's $\rho$ & \href{https://huggingface.co/datasets/tyang816/TAPE_Stability}{\texttt{tyang816/TAPE\_Stability}} \\
    TAPE\_Fluorescence (FLU) & - & - & $21,446$ & $5,362$ & $27,217$ & Spearman's $\rho$ & \href{https://huggingface.co/datasets/tyang816/TAPE_Fluorescence}{\texttt{tyang816/TAPE\_Fluorescence}} \\
    \midrule
    \multicolumn{8}{c}{\textbf{Other}} \\
    \midrule
    MetalIonBinding (MIB) & $92.36_{(6.43)}$ & $83.66_{(8.73)}$ & $5,068$ & $662$ & $665$ & accuracy & \href{https://huggingface.co/datasets/tyang816/MetalIonBinding}{\texttt{tyang816/MetalIonBinding}} \\
    Thermostability (TMO) & $79.02_{(12.26)}$ & $74.60_{(13.82)}$ & $5,054$ & $639$ & $1,336$ & Spearman's $\rho$ & \href{https://huggingface.co/datasets/tyang816/Thermostability}{\texttt{tyang816/Thermostability}} \\
    DeepET\_Topt (DET) & $92.98_{(5.32)}$ & $85.18_{(8.74)}$ & $1,478$ & $185$ & $185$ & Spearman's $\rho$ & \href{https://huggingface.co/datasets/tyang816/DeepET_Topt}{\texttt{tyang816/DeepET\_Topt}} \\
    SortingSignal (SIG) & $81.09_{(11.66)}$ & - & $1,484$ & $185$ & $186$ & f1\_max & \href{https://huggingface.co/datasets/tyang816/SortingSignal}{\texttt{tyang816/SortingSignal}} \\ \bottomrule
    \end{tabular}
}
\caption{Overview of the selected datasets for evaluating, including localization, solubility, annotation, mutation effects, and other properties. The table lists dataset sizes, evaluation metrics, and pLDDT from \texttt{AlphaFold2} and \texttt{ESMFold}, with standard deviations in parentheses.}\label{tab:dataset}
\end{center}
\end{table*}
\section{Evaluated Benchmark Datasets}\label{app:benchmark}
Table \ref{tab:dataset} summarizes datasets used for training and evaluating PLMs. The columns provide details on training, validation, and test splits, evaluation metrics (\eg, accuracy, F1-score, Spearman’s correlation), and implementation sources. Additionally, the mean and standard deviation of \texttt{AlphaFold2} (AF2) and \texttt{ESMFold} (EF) predicted confidence scores (pLDDT) are reported. For \textbf{FLIP\_AAV} and \textbf{FLIP\_GF1}, we only selected the sampled partitioning method for testing.

\begin{table}[ht]
\resizebox{\linewidth}{!}{
    \begin{tabular}{@{}lll@{}}
    \toprule
    \textbf{Short Name} & \textbf{Metrics Name} & \textbf{Problem Type} \\ \midrule
    accuracy & Accuracy & single/multi-label cls \\
    recall & Recall & single/multi-label cls \\
    precision & Precision & single/multi-label cls \\
    f1 & F1Score & single/multi-label cls \\
    mcc & MatthewsCorrCoef & single/multi-label cls \\
    auc & AUROC & single/multi-label cls \\
    f1\_max & F1ScoreMax & multi-label cls \\
    spearman\_corr & SpearmanCorrCoef & regression \\
    mse & MeanSquaredError & regression \\ \bottomrule
    \end{tabular}
}
\caption{Supported metrics with abbreviations. "Single-label cls" refers to single-label classification tasks, while "multi-label cls" refers to classification tasks where multiple labels can be assigned to each instance.}
\label{tab:metrics}
\end{table}

\section{Metrics}\label{app:metrics}
Table \ref{tab:metrics} lists the supported evaluation metrics, abbreviations, and corresponding problem types.

\section{Collection}\label{app:collection}
\subsection{Introduction}
\textbf{Collection} is designed for automated extraction of protein-related data from \texttt{InterPro}, \texttt{RCSB PDB}, \texttt{UniProt}, and \texttt{AlphaFold DB}. It supports structured metadata, sequence information, and 3D structural data retrieval, streamlining large-scale protein engineering research\footnote{\url{https://github.com/tyang816/VenusFactory/blob/main/download/README.md}}.
\subsection{Implementation and Workflow}
Implemented in Python, \textbf{Collection} leverages \texttt{requests} for API interactions and multiprocessing for parallel processing. It supports both single and batch retrieval via text or \texttt{JSON} input. The workflow consists of input parsing, data fetching, data processing, and file storage, with structured output in \texttt{FASTA}, \texttt{JSON}, \texttt{PDB}, and \texttt{mmCIF} formats. API requests include error handling with automatic retries to manage rate limits and network failures.
\subsection{Data Organization}
Output is stored hierarchically, with metadata, sequences, and structures categorized for easy access. For instance, \texttt{InterPro} metadata includes domain details (detail.json), accession metadata (meta.json), and associated \texttt{UniProt} IDs (uids.txt). UniProt sequences are saved in \texttt{FASTA} format, with an option to merge entries, while AlphaFold structures are organized by ID prefix for optimized storage.
\subsection{Error Handling and Logging}
\textbf{Collection} logs failed downloads in "failed.txt", recording network timeouts, missing IDs, and API errors for debugging and reattempts. Parallel downloading, caching, and adaptive rate limiting enhance retrieval efficiency, reducing redundant API calls and optimizing request frequency.

\end{document}